\pdfoutput=1
\documentclass[11pt]{article}

\usepackage[final]{acl}

\usepackage{times}
\usepackage{latexsym}

\usepackage[T1]{fontenc}
\usepackage[utf8]{inputenc}

\usepackage{microtype}

\usepackage{inconsolata}

\usepackage{graphicx}
\usepackage{amsmath}
\usepackage{amsfonts}
\newcommand{\argmax}{\operatorname*{arg\,max}}
\usepackage{algorithm}
\usepackage{algpseudocode}
\usepackage{tabularray}
\usepackage{xcolor,colortbl}
\usepackage{textcomp}
\usepackage{multicol}
\usepackage{subcaption}
\usepackage{booktabs}
\usepackage{arydshln}
\usepackage{xspace}
\usepackage{array}
\newcommand\sect[1]{\S\ref{#1}}
\definecolor{Gray}{gray}{0.93}

\newcommand{\ie}[0]{\emph{i.e., }}

\newcommand{\eg}[0]{\emph{e.g., }}

\newcommand\ours{\textsc{RiOT}\xspace}
\definecolor{gred}{HTML}{cc0200}
\definecolor{ggreen}{HTML}{4C9F26}
\newcommand{\uaa}[1]{\scriptsize\textcolor{ggreen}{\footnotesize $\uparrow$}{\color{ggreen}#1}}
\newcommand{\daa}[1]{\scriptsize\textcolor{gred}{\footnotesize $\downarrow$}{\color{gred}#1}}

\title{\ours: Efficient Prompt Refinement with Residual Optimization Tree }

\author{
 \textbf{Chenyi Zhou\textsuperscript{1}},
 \textbf{Zhengyan Shi\textsuperscript{2}}, \\
 \textbf{Yuan Yao\textsuperscript{1}},
 \textbf{Lei Liang\textsuperscript{3}},
 \textbf{Huajun Chen\textsuperscript{1}},
 \textbf{Qiang Zhang\textsuperscript{1†}}
\\
\\
 \textsuperscript{1}Zhejiang University,
 \textsuperscript{2}University College London,
 \textsuperscript{3}Ant Group \\
}

\begin{document}
\maketitle
\let\thefootnote\relax\footnote{† Corresponding author: qiang.zhang.cs@zju.edu.cn}

\begin{abstract}
Recent advancements in large language models (LLMs) have highlighted their potential across a variety of tasks, but their performance still heavily relies on the design of effective prompts. Existing methods for automatic prompt optimization face two challenges: lack of diversity, limiting the exploration of valuable and innovative directions and semantic drift, where optimizations for one task can degrade performance in others. To address these issues, we propose Residual Optimization Tree (\ours), a novel framework for automatic prompt optimization. \ours iteratively refines prompts through text gradients, generating multiple semantically diverse candidates at each step, and selects the best prompt using perplexity. Additionally, \ours incorporates the text residual connection to mitigate semantic drift by selectively retaining beneficial content across optimization iterations. A tree structure efficiently manages the optimization process, ensuring scalability and flexibility. Extensive experiments across five benchmarks — covering commonsense, mathematical, logical, temporal, and semantic reasoning — demonstrate that \ours outperforms both previous prompt optimization methods and manual prompting. Our code is released at \href{https://github.com/Qing1Zhong/RiOT}{\texttt{https://github.com/Qing1Zhong/RiOT}}.
\end{abstract}

\section{Introduction}
Recent advances in large language models (LLMs) have showcased their exceptional performance across a wide range of tasks~\cite{bommasani2021opportunities, li2022competition, gpt4_report, gemini1.5_report,Chen_LLaMA,zhang_agent,jiang-etal-2024-med,XU2025126585,bai2025qwen2,jiang2025hscr}. However, the effectiveness of these models often relies heavily on the careful design of prompts~\cite{brown2020language, reynolds2021prompt, kojima2022large, chain-of-thought, amatriain2024prompt,Chen_Perspectives,jiang2024joint,jiang2025fast}, typically involving labor-intensive trial-and-error processes that require substantial domain expertise and computational resources. Consequently, there is an increasing demand for methods that can automatically optimize prompts, simplifying the task of designing high-performance prompts.

\begin{figure*}[!ht]
    \centering
    \includegraphics[width=1\linewidth]{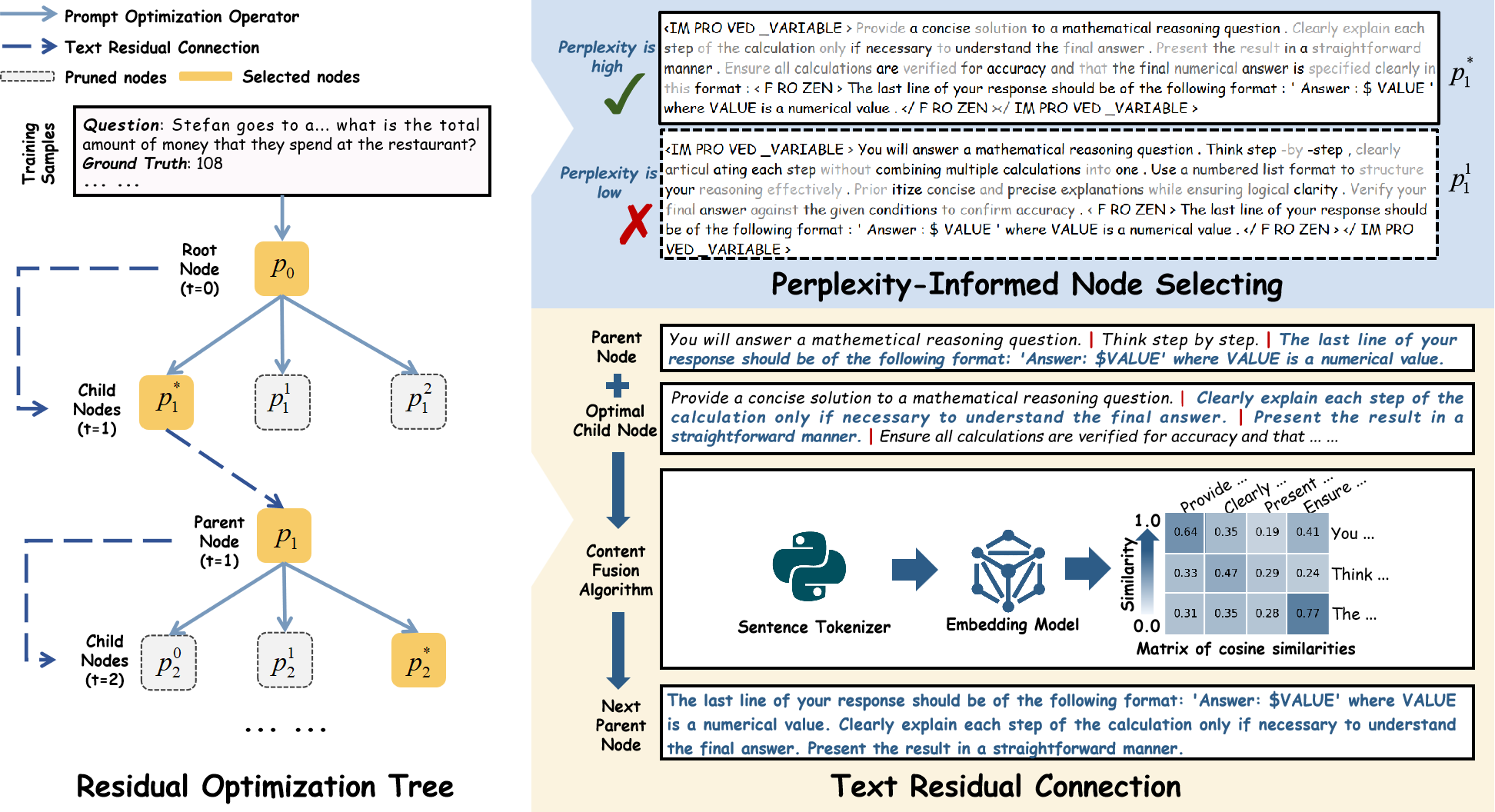}
    \caption{Overview of \ours. \textbf{(i)} At each step, \ours generates multiple candidate prompts (child nodes). \textbf{(ii)} Then, the optimal child node is selected based on perplexity. The perplexity-informed selection process is shown in the region within the blue block \textcolor[rgb]{0.51,0.67,0.86}{\rule{5mm}{2mm}}. \textbf{(iii)} Finally, the parent node is connected to the optimal child node based on semantic similarity. This residual connection process is highlighted within the yellow block \textcolor[rgb]{0.97,0.88,0.65}{\rule{5mm}{2mm}}.}
    \label{main figure}
\end{figure*}

Recently, researchers have begun to explore automatic prompt optimization by directly leveraging LLM outputs and token logits.
\citet{APE} first introduce Automatic Prompt Engineer (APE) concept, showing that LLMs can self-generate effective prompts given a small set of input-output pair demonstrations. Subsequent studies~\cite{APO, OPRO, textgrad} expanded on this by integrating concepts such as optimizers and text gradients, using meta-prompts to guide LLMs in refining prompts iteratively based on feedback from training samples.

Despite the promising performance of existing methods~\cite{APE, APO, OPRO, textgrad}, they face two challenges. First, these approaches often struggle with limited diversity during the optimization process. For example, \citet{OPRO, textgrad, APO} generate only one candidate prompt per iteration, restricting the breadth of the search space. While \citet{APE} generates multiple candidates in parallel, it lacks mechanisms for iterative refinement, limiting its ability to optimize the generated prompts effectively. Second, iterative prompt optimization can lead to semantic drift. This occurs when optimizing a prompt for one task inadvertently degrades its performance in other, previously successful scenarios. This issue is similar to the stability-plasticity dilemma observed in continual learning~\cite{mccloskey1989catastrophic, catastrophic_forgetting, ren2024analyzing}, but has yet to be thoroughly explored in the context of prompt optimization in discrete combinatorial spaces.

To address these challenges, we propose Residual Optimization Tree (\ours), a novel framework for automatic prompt optimization (see \sect{sec:method}). Building on previous works~\cite{APE, OPRO, textgrad}, \ours iteratively refines prompts using text gradients. However, unlike prior methods, \ours generates multiple candidate prompts with distinct semantic meanings at each optimization step.
Next, \ours selects the best candidate by calculating the perplexity of each prompt. This approach substantially enhances the diversity during the optimization process while reducing additional computational overhead. To mitigate the risk of semantic drift, \ours introduces a text residual connection algorithm, which selectively preserves the content of candidates from different optimization iterations based on semantic similarity (see \sect{sec:residual learning}).
Additionally, \ours employs a tree structure to efficiently manage the hierarchical nature of the optimization process. Each level in the tree corresponds to an optimization iteration, with each node representing a candidate prompt. The selection process acts as a pruning mechanism, while text residual connections link nodes across different levels, ensuring that beneficial information is retained. This tree structure not only enables effective tracking of the optimization process but also guarantees scalability and flexibility.

\ours demonstrates robust performance in prompt optimization, substantially enhancing the capabilities of LLMs. Across five diverse benchmarks—spanning commonsense reasoning, mathematical reasoning, logical reasoning, temporal understanding, and semantic understanding (see \sect{sec:setup})—\ours outperforms both previous prompt optimization methods and manual prompting (see \sect{sec:results}). For example, on the GSM8K benchmark~\cite{GSM8K}, \ours improves accuracy by 4.6\% (in absolute) over the leading baseline.
Our contributions can be summarized as follows: 
\begin{itemize} 
\item We introduce the first tree-based framework with a residual connection for automatic prompt optimization. 
\item We address the challenge of limited diversity in prompt optimization by effectively exploring prompt space.
\item We mitigate the phenomenon of semantic drift by introducing residual learning that retains crucial elements in optimization.
\end{itemize}

\section{Related Work}
\paragraph{Prompt Optimization.} The majority of work on prompt optimization can be categorized into two main types. The first category focuses on soft prompt tuning, where prompts are represented as task-specific continuous vectors and trained through gradient-based methods~\cite{lester-etal-2021-power,li-liang-2021-prefix,qin-eisner-2021-learning,liu2024gpt}. The second category employs discrete token search, using gradient-guided approaches~\cite{shin-etal-2020-autoprompt,gao-etal-2021-making,shi-etal-2023-toward,wen2024hard} or reinforcement learning~\cite{deng-etal-2022-rlprompt,zhang2023tempera}. These works require full access to the language model's parameters or auxiliary reward models, which face substantial limitations when applied to modern LLMs accessible only via API. This underscores the need for black-box strategies that rely solely on textual feedback to optimize the prompt systematically. 
Some recent works have developed gradient-free methods for prompt optimization. This approach was proposed by \citet{APE}, which involved instructing LLMs to infer the prompt. \citet{APO} later formalized the concept of text gradient to refer to textual feedback-based optimization. Subsequent works have further improved upon this by incorporating optimization logic in the prompt space~\cite{OPRO,textgrad}. However, these methods are limited by a constrained optimization space, which hampers diversity, and fail to address the issue of semantic drift, resulting in suboptimal performance in dynamic contexts.                

\paragraph{Semantic Drift.} Catastrophic forgetting typically occurs in the context of continual learning. In such scenarios, neural networks may lose previously acquired knowledge when adapting to new tasks, a challenge that has inspired prior work to stabilize parameter updates in continuous spaces through methods such as parameter isolation and regularization~\cite{li2017learning,kirkpatrick2017overcoming,maltoni2019continuous,dou-etal-2024-loramoe}. The core of these approaches lies in preserving the original weights as much as possible during updates. A similar issue arises in discrete prompt optimization, where iterative modification of prompts can overwrite crucial semantic components, this phenomenon we refer to as semantic drift. In this paper, we address this issue by dynamically preserving high-value semantic components in the prompts through text residual connection.

\begin{algorithm*}[t]
\caption{Content Fusion Algorithm}
\label{algorithm}
\begin{algorithmic}[1]
\Statex \hspace{-1.4em} \textbf{Require:} Step $t$, parent node $\mathcal{P}_{t-1}$, the optimal child node $\mathcal{P}_{t}^*$, a pretrained embedding model $\mathcal{E}(\cdot)$, a sentence tokenizer $\mathcal{T}(\cdot)$, and two hyperparameters $b_1$ and $b_2$.
\Statex \hspace{-1.4em} \textbf{Divide into sentences:} 
\State $\{s_0, s_1, \dots, s_n\} \gets \mathcal{T}(\mathcal{P}_{t-1})$, $\{r_0, r_1, \dots, r_m\} \gets \mathcal{T}(\mathcal{P}_{t}^*)$ \Comment{Tokenize $\mathcal{P}_{t-1}$ and $\mathcal{P}_{t}^*$ using $\mathcal{T}$}
\Statex \hspace{-1.4em} \textbf{Obtain the semantic representation:} 
\State $\mathbf{E}_{t-1} \gets \{\mathcal{E}(s_i) \mid i \in [1, n]\}$, $\mathbf{E}_{t}^* \gets \{\mathcal{E}(r_j) \mid j \in [1, m]\}$ \Comment{Embed the sentences using $\mathcal{E}$}. $\mathbf{E}_{t-1}$ is an $n \times d$ matrix of embeddings for $\mathcal{P}_{t-1}$. $\mathbf{E}_{t}^*$ is an $m \times d$ matrix of embeddings for $\mathcal{P}_{t}^*$. $d$ is the embedding dimension.
\Statex \hspace{-1.4em} \textbf{compute cosine similarities:} 
\State $\textbf{sim}_{ab} \gets \frac{\mathbf{E}_{t-1} \cdot \mathbf{E}_{t}^{*-T}}{\|\mathbf{E}_{t-1}\|_2 \|\mathbf{E}_{t}^*\|_2}$ \Comment{Matrix of cosine similarities of size $n \times m$}
\Statex \hspace{-1.4em} \textbf{Select sentences based on similarity thresholds:} 
\State $selected\_t \gets \{s_{idx} \mid \max(\text{sim}_{ab}[idx, :]) \geq 1 - b_1 \}$ \Comment{Select sentences from $\mathcal{P}_{t-1}$}
\State $selected\_r \gets \{r_{idx} \mid \max(\text{sim}_{ab}[:, idx]) < 1 - b_2 \}$ \Comment{Select sentences from $\mathcal{P}_{t}^*$}
\State $\mathcal{P}_t \gets selected\_t \cup selected\_r$
\State \textbf{Return} $\mathcal{P}_t$
\end{algorithmic}
\end{algorithm*}

\section{Preliminaries}
\subsection{Problem Definition}
We define the task dataset as $\mathcal{D} = \{(x_i, y_i)\}_{i=1}^{N}$, where $x_i$ represents the natural language query, $y_i$ denotes the corresponding ground truth, and $N$ is the total number of test samples. The objective of the prompt optimization task is to search for the optimal prompt $p^*$ that maximizes the LLM performance $\mathcal{F}(\cdot)$ on the given task. This task can be formally defined as:
\begin{equation}
    p^* = \argmax_{p \in \mathcal{P}_{\text{space}}} \sum_{i=1}^{N} \mathcal{S}\left( \mathcal{F}(x_i; p), y_i \right),
\end{equation}
where $\mathcal{P}_{\text{space}}$ represents the set of all possible prompts (\eg manually designed prompts or automatically generated prompts) and $\mathcal{S}(\cdot)$ denotes the corresponding evaluation metrics. 

\subsection{Prompt Optimization Operator}
Several frameworks for automatic prompt optimization typically employ similar optimization logic, achieving improved performance. We select TextGrad~\cite{textgrad} as the backbone for executing optimization at each step, as it both follows this successful path and offers great extensibility.
For clarity, we define the model responsible for evaluating prompt performance as the target model $\mathcal{F}_\text{target}(\cdot)$, and the model used for feedback generation or prompt optimization as the optimization model $\mathcal{F}_\text{opt}(\cdot)$.
Let $\mathcal{D}_\text{train}$ denote the training set. At step $t$, the target model $\mathcal{F}_\text{target}(\cdot)$ generates responses $\{\hat{y}_{i}^{(t)}\}$ using the current prompt $p_t$ and computes the loss $\mathcal{L}(p_t)$ over $\mathcal{D}_\text{train}$: 
\begin{equation}
    \{\hat{y}_i^{(t)}\} = \mathcal{F}_\text{target}(\mathcal{D}_\text{train}, p_t)
\end{equation}
\begin{equation}
    \mathcal{L}(p_t) = \mathcal{S}(\mathcal{D}_\text{train}, \{\hat{y}_{i}^{(t)}\}).    
\end{equation}
Then, the optimizer model proposes the candidate prompt $p_{t+1}$:
\begin{equation}
    p_{t+1} = \mathcal{F}_\text{opt}(\nabla_{p_t}\mathcal{L}(p_t), p_t).
\end{equation}
where $\nabla_{p_t}\mathcal{L}(p_t)$ denotes the text gradient encoding improvement directions in natural language. We encapsulate this iterative process into a Prompt Optimization Operator $\mathcal{M}(\cdot)$, defined as:
\begin{equation}
    p_{t+1} = \mathcal{M}(\mathcal{D}_\text{train},p_t)
\end{equation}
This operator abstraction enables modular integration with downstream components.

\section{Residual Optimization Tree}
\label{sec:method}
In this section, we introduce our proposed method Residual Optimization Tree (\ours), as shown in Figure~\ref{main figure}. Specifically, \ours starts with an initial prompt $p_0$ as the root node. At step $t$, each parent node $p_t$ produces $K$ candidate child nodes $\{p_{t+1}^{(i)}\}_{i=1}^K$ through prompt optimization operator $\mathcal{M}(\cdot)$ (\ie the width of tree is $K$). Since LLMs exhibit inherent variability in their outputs, the same input may yield different outputs. To account for this, \ours computes the loss $\mathcal{L}(p_t)$ $K$ times, generating $K$ different candidate prompts. {Then we propose a dynamic pruning strategy to select the optimal child node $p_{t+1}^*$ from candidates $\{p_{t+1}^{(i)}\}_{i=1}^K$ (explained in \sect{sec:entropy}). This approach fosters greater diversity within the prompt space without sacrificing quality.}

To address the challenge of semantic drift, we introduce the Text Residual Connection (detailed in \sect{sec:residual learning}). This mechanism is inspired by the concept of residual learning in deep networks and ensures that semantic components from the parent node can effectively be transferred to child nodes during iterative optimization, reducing excessive divergence between parent and child nodes.

\subsection{Perplexity-Informed Node Selection}
\label{sec:entropy}
A common strategy of child node selection often relies solely on the quality of the candidate nodes generated by the LLM, overlooking the importance of diversity in the optimization process. This can result in less innovative outcomes, limiting the exploration of valuable prompt combinations and missing innovative directions. Instead, we introduce a perplexity-informed approach to select the most promising child node. The idea is to prioritize nodes with higher semantic diversity.

For each child node candidate $p_{t+1}^{(i)}$, we compute its perplexity $PPL(p_{t+1}^{(i)})$, which measures the uncertainty of the model's prediction for the candidate content. Given the candidate prompt $p_{t+1}^{(i)}=(x_0,x_1,\cdots,x_J)$, the perplexity $PPL(p_{t+1}^{(i)})$ is defined as:
\begin{equation}
    PPL(p_{t+1}^{(i)}) = \exp\{-\frac{1}{J}\sum_{j=1}^{J}p_{\theta}(x_j \mid x_{<j})\},
\end{equation}
where $p_{\theta}(x_j \mid x_{<j})$ is the likelihood of the $j$-th token conditioned on the preceding tokens $x_{<j}$ according to the LLM. The perplexity value is then incorporated into the selection process. First, from an information theoretic perspective~\cite{shannon1948mathematical}, higher perplexity indicates lower token co-occurrence probability, suggesting more information. Second, established practices in Bayesian optimization~\cite{snoek2012practical} preferentially explore regions of high uncertainty to maximize information gain. Therefore, we select the optimal child node $p_{t+1}^*$ by maximizing the perplexity value:
\begin{equation}
    p_{t+1}^* = \arg \max_{i} PPL(p_{t+1}^{(i)}).
\end{equation}
This perplexity-informed node selection approach enhances \ours's ability to traverse a diverse and effective prompt space.

\subsection{Text Residual Connection}
\label{sec:residual learning}
To mitigate semantic drift during the iterative prompt optimization process, we propose the Text Residual Connection, which is inspired by residual learning techniques in deep neural networks. The core idea is to ensure that the important information from the parent prompt is preserved and effectively transferred to the successor prompt, preventing the forgetting phenomenon that can occur when iteratively modifying prompts. In traditional neural networks \cite{he2016deep}, residual connections facilitate the learning of differences between the input and target representations, enabling deeper networks without suffering from vanishing gradients. {We adopt a similar concept to prompt optimization. Initializing with prompt $p_{0}$ as the tree root, at step $t$ \ours generate an optimized child node $p_t^*$ from the parent node $p_{t-1}$. Contrary to previous strategies that directly propagate $p_t^*$ as the immediate successor $p_t$, \ours computes the semantic residual components which represent the difference between the child node $p_t^*$ and the parent node $p_{t-1}$, then fuses them through controlled composition, as follows:
\begin{equation}
    p_t = \mathcal{G}(p_{t-1}, p_t^*).
\end{equation}
Here, $\mathcal{G}(\cdot)$ represents the \textbf{Content Fusion Algorithm} outlined in Algorithm~\ref{algorithm}, which is achieved by adjusting two hyperparameters, $b_1$ and $b_2$, that control the degree of fusion between the parent and child prompts. This ensures that the successor prompt $p_t$ retains the meaningful elements from the parent that contribute to the optimization trajectory, while introducing new elements from the optimized child node.

In essence, the Text Residual Connection can be viewed as a way of progressively refining the prompt, ensuring that each iteration of optimization builds upon the information already captured by the previous node, rather than starting from scratch, thereby semantic consistency across all prompts in the optimization tree.

\section{Experimental Step} 
\label{sec:setup}
\paragraph{Datasets.}
We evaluate \ours across five benchmarks, including \textbf{LogiQA 2.0}~\cite{logiqa}, \textbf{StrategyQA}~\cite{StrategyQA}, \textbf{Object Counting}~\cite{Big-Bench, BBH}, \textbf{GSM8K}~\cite{GSM8K}, and \textbf{Date Understanding}~\cite{Big-Bench, BBH}, which are designed to cover diverse task types such as temporal understanding, semantic reasoning, mathematical computation, and commonsense inference. For detailed dataset distribution, please refer to the Appendix~\ref{appendix: dataset}.

\begin{table*}[ht]
\centering
\resizebox{\textwidth}{!}{
\begin{tabular}{lccccc}
\toprule
           & \multicolumn{2}{c}{\bf \textsc{True / False}} & \multicolumn{2}{c}{\bf \textsc{Generative}} & \bf \textsc{Multiple-choice} \\
           \cmidrule(lr){2-3}                          \cmidrule(lr){4-5}                        \cmidrule(lr){6-6}
\bf Method & LogiQA 2.0 & StrategyQA & Object Counting & GSM8K & Date Understanding \\
           & (N=160) & (N=100) & (N=210) & (N=100) & (N=329) \\
\midrule
\rowcolor{Gray} \multicolumn{6}{l}{\textit{*Manual Prompting Methods}} \\
Zero-Shot CoT 
                & 59.0 \textcolor{gray}{\tiny \textpm 1.5}     
                & 65.8 \textcolor{gray}{\tiny \textpm 1.9}     
                & 71.0 \textcolor{gray}{\tiny \textpm 1.9}
                & 60.2 \textcolor{gray}{\tiny \textpm 1.5}      
                & 76.1 \textcolor{gray}{\tiny \textpm 1.0}
                \\
Four-Shot CoT   & 59.7 \textcolor{gray}{\tiny \textpm 3.5} \uaa{0.7}
                & 71.2 \textcolor{gray}{\tiny \textpm 1.0} \uaa{5.4}
                & 71.2 \textcolor{gray}{\tiny \textpm 2.2} \uaa{0.2}
                & 73.4 \textcolor{gray}{\tiny \textpm 1.5} \uaa{13.2}
                & 73.2 \textcolor{gray}{\tiny \textpm 1.7} \daa{2.9}
                \\
Twenty-Shot CoT 
                & 59.4 \textcolor{gray}{\tiny \textpm 1.0} \uaa{0.4}             
                & 71.6 \textcolor{gray}{\tiny \textpm 0.8} \uaa{5.8}           
                & 70.8 \textcolor{gray}{\tiny \textpm 1.4} \daa{0.2}              
                & 72.8 \textcolor{gray}{\tiny \textpm 1.2} \uaa{12.6}          
                & \underline{76.3} \textcolor{gray}{\tiny \textpm 1.9} \uaa{0.2}             
                \\ 
\midrule
\rowcolor{Gray} \multicolumn{6}{l}{\textit{*Automatic Prompt Optimization Methods}} \\
APE~\cite{APE} 
                & 57.4 \textcolor{gray}{\tiny \textpm 3.5} \daa{1.6}
                & 70.4 \textcolor{gray}{\tiny \textpm 3.3} \uaa{4.6}
                & 79.2 \textcolor{gray}{\tiny \textpm 1.9}  \uaa{8.2}
                & 76.6 \textcolor{gray}{\tiny \textpm 1.9} \uaa{16.4}
                & 72.8 \textcolor{gray}{\tiny \textpm 2.8} \daa{3.3}
                \\
OPRO~\cite{OPRO}            
                & 58.0 \textcolor{gray}{\tiny \textpm 4.8} \daa{1.0}   
                & 67.0 \textcolor{gray}{\tiny \textpm 3.0} \uaa{1.2}          
                & 79.9 \textcolor{gray}{\tiny \textpm 1.3} \uaa{8.9}               
                & 72.8 \textcolor{gray}{\tiny \textpm 2.0} \uaa{12.6}          
                & \underline{76.3} \textcolor{gray}{\tiny \textpm 0.5} \uaa{0.2}              
                \\
TextGrad~\cite{textgrad}        
                & \underline{60.0} \textcolor{gray}{\tiny \textpm 1.2} \uaa{1.0}
                & 68.8 \textcolor{gray}{\tiny \textpm 1.7} \uaa{3.0}  
                & \textbf{88.3} \textcolor{gray}{\tiny \textpm 1.2} \uaa{17.3}      
                & 74.8 \textcolor{gray}{\tiny \textpm 1.0} \uaa{14.6}    
                & 71.9 \textcolor{gray}{\tiny \textpm 0.7} \daa{4.2}               
                \\
DSPy~\cite{DSPy}
        & 59.8 \textcolor{gray}{\tiny \textpm 1.6} \uaa{0.8} 
        & \underline{73.4} \textcolor{gray}{\tiny \textpm 1.5} \uaa{7.6}
        & 84.5 \textcolor{gray}{\tiny \textpm 2.2} \uaa{13.5}
        & \underline{79.0} \textcolor{gray}{\tiny \textpm 1.7} \uaa{18.8}
        & 74.3 \textcolor{gray}{\tiny \textpm 1.1} \daa{1.8}{}\\
\hdashline\noalign{\vskip 0.4ex}
\ours (\textbf{Ours}) & \textbf{61.4} \textcolor{gray}{\tiny \textpm 1.5} \uaa{2.4}
                & \textbf{74.6} \textcolor{gray}{\tiny \textpm 1.5} \uaa{8.8}       
                & \underline{86.9} \textcolor{gray}{\tiny \textpm 0.6} \uaa{15.9}            
                & \textbf{81.2} \textcolor{gray}{\tiny \textpm 1.2} \uaa{21.0}          
                & \textbf{78.2} \textcolor{gray}{\tiny \textpm 0.6}  \uaa{2.1}
                \\
\bottomrule
\end{tabular}
}
\vspace{-0.5em}
\caption{Comparison of Automatic Prompt Optimization Methods. We report mean accuracy (\%) and standard deviation. \textbf{Bold} and \underline{underlined} values indicate best and second-best performance, respectively. Red arrows indicate the absolute performance decrease, while green arrows indicate the absolute performance increase, both compared to the Zero-Shot CoT prompting. $N$ represents the number of samples used for testing.}
\label{main result}
\end{table*}

\paragraph{Baselines.}
We compare \ours to prior works in two categories: 
\begin{itemize}
\item \textbf{Manual Prompting Methods.} We experiment with (1) zero-shot Chain-of-Thought (CoT) prompting and (2) few-shot CoT prompting~\cite{kojima2022large,chain-of-thought}. See Appendix~\ref{appendix: manual prompt} for the prompt template construction details.

\item \textbf{Automatic Prompt Optimization Methods.} We compare with four established approaches. (1) APE~\cite{APE} leverages LLMs to generate candidate prompts based on task-specific input-output pairs, followed by a selection process to identify the optimal prompts. (2) OPRO~\cite{OPRO} positions language models as optimizers, utilizing the meta-prompt to systematically refine the initial prompt. (3) TextGrad~\cite{textgrad} adopts a parameterization perspective by treating prompts as optimizable parameters while introducing the concept of textual gradients for optimization. {(4) DSPy~\cite{DSPy} formulates prompt optimization as a declarative program synthesis task.}
\end{itemize}

\paragraph{Implementation Details.}
Following prior work~\cite{textgrad}, we configure both GPT-3.5-turbo~\cite{GPT3.5-turbo} (as the target model) and GPT-4o~\cite{GPT4o} (as the optimization model) with temperature parameters fixed at 0. Moreover, we tokenize the prompt using the NLTK sentence tokenizer~\cite{nltk}. The \texttt{text-embedding-3-large}  model~\cite{openai2023textembedding} is used for embedding, with $b_1$ set to $0.25$ and $b_2$ set to $0.5$. $K$ is set to $3$. A batch size of $4$ is utilized with $3$ training epochs, yielding $15$ iterations through the optimization process. We use the zero-shot CoT prompts as initial prompts for prompt optimization methods.   

We report accuracy with corresponding standard deviation, based on five independent runs using fixed prompts and test samples. This approach accounts primarily for the variance introduced by the decoding strategy. The optimal prompt, identified on the validation set, is used for testing.

\begin{figure*}
\centering
\includegraphics[width=\textwidth]{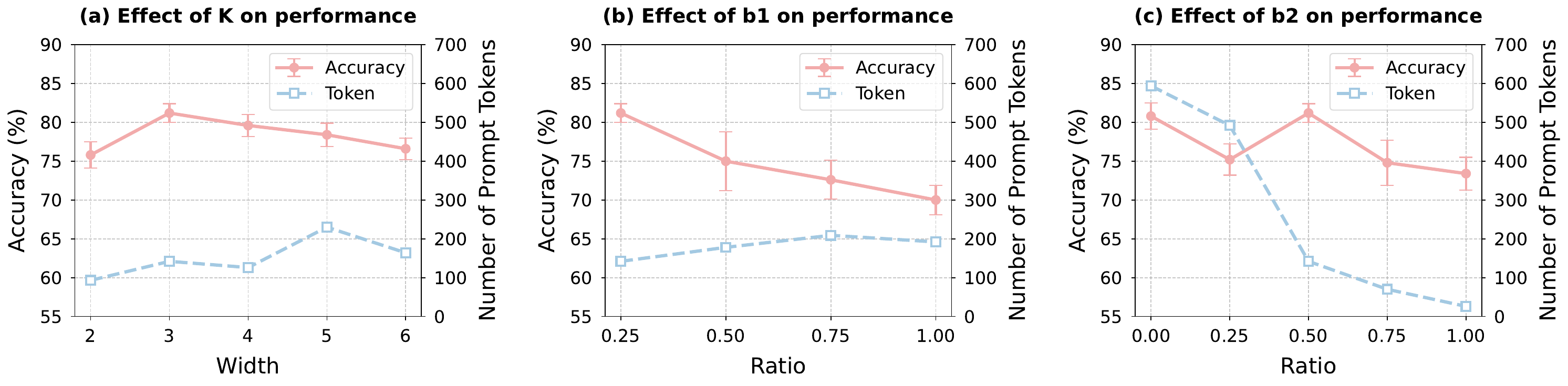}
\vspace{-2em}
\caption{Hyperparametric Senstivity Analysis of \ours. 
Our results show that
\textbf{(a)}: Increasing the tree width $K$ leads to a unimodal curve, with peak accuracy at $K=3$; 
\textbf{(b)}: The parameter $b_1$ is negatively correlated with performance; and
\textbf{(c)}: The parameter $b_2$ plays a critical role in balancing performance and computational overhead.}
\vspace{-0.5em}
\label{figure:performance}
\end{figure*}

\section{Results}
\label{sec:results}
\subsection{Main Results}

\paragraph{Finding \#1: Scaling the number of demonstration examples in few-shot CoT prompting leads to diminishing returns.}
As shown in Table~\ref{main result}, while few-shot CoT outperforms zero-shot across tasks, scaling the number of demonstrations yields minimal gains. For instance, on StrategyQA, expanding the prompt from 4 to 20 examples improves accuracy by merely 0.4\%, while on LogiQA 2.0, Twenty-shot CoT prompting unexpectedly underperforms Four-shot CoT prompting by 0.3\%. We posit that this phenomenon stems from two critical factors: (1) the escalating prompt length challenges models' contextual comprehension due to attention dilution, and (2) redundant examples often fail to provide task-specific inductive signals, forcing models to implicitly extrapolate patterns from noisy demonstrations. 
\begin{table*}[ht]
\centering
\resizebox{\textwidth}{!}{
\begin{tabular}{lccccc}
\toprule
           & \multicolumn{2}{c}{\bf \textsc{True / False}} & \multicolumn{2}{c}{\bf \textsc{Generative}} & \bf \textsc{Multiple-choice} \\
           \cmidrule(lr){2-3}                          \cmidrule(lr){4-5}                        \cmidrule(lr){6-6}
\bf Experiment & LogiQA 2.0 & StrategyQA & Object Counting & GSM8K & Date Understanding \\
           & (N=160) & (N=100) & (N=210) & (N=100) & (N=329) \\
\midrule
Baseline 
                & 65.9 \textcolor{gray}{\tiny \textpm 0.8}          
                & 79.6 \textcolor{gray}{\tiny \textpm 1.9}       
                & \underline{92.6} \textcolor{gray}{\tiny \textpm 0.9}
                & 93.4 \textcolor{gray}{\tiny \textpm 0.8}      
                & \underline{89.9} \textcolor{gray}{\tiny \textpm 0.4}               \\
Prompt Transferability
                & \underline{67.0} \textcolor{gray}{\tiny \textpm 0.5} \uaa{1.1}
                & \underline{81.4} \textcolor{gray}{\tiny \textpm 1.0} \uaa{1.8}
                & 90.2 \textcolor{gray}{\tiny \textpm 0.9} \daa{2.4}
                & \underline{93.6} \textcolor{gray}{\tiny \textpm 0.8} \uaa{0.2}
                & 86.7 \textcolor{gray}{\tiny \textpm 0.3} \daa{3.2}
                \\
Model Transferability            
                & \textbf{69.4} \textcolor{gray}{\tiny \textpm 0.4} \uaa{3.5}           
                & \textbf{82.0} \textcolor{gray}{\tiny \textpm 0.6} \uaa{2.4}           
                & \textbf{93.0} \textcolor{gray}{\tiny \textpm 0.6} \uaa{0.4}           
                & \textbf{95.0} \textcolor{gray}{\tiny \textpm 0.6} \uaa{1.6}          
                & \textbf{90.5} \textcolor{gray}{\tiny \textpm 0.6} \uaa{0.6}
                \\
\bottomrule
\end{tabular}
}
\caption{Generalization performance of \ours across five datasets. All experiments are evaluated on Gemini-1.5-flash, and optimizer model is GPT-4o. \textbf{Baseline}: Applying the zero-shot CoT prompt. \textbf{Prompt Transferability}: Prompts are optimized for GPT-3.5-turbo. \textbf{Model Transferability}: Prompts are optimized for Gemini-1.5-flash. 
\textbf{Bold} and \underline{underlined} values indicate best and second-best performance, respectively.
Red arrows indicate the absolute performance decrease, while gree arrows indicate the absolute performance increase, both compared to the baseline.
$N$ represents the number of samples used for testing.
}
\vspace{-0.6em}
\label{generalization}
\end{table*}

\paragraph{Finding \#2: \ours outperforms manual prompting and automatic prompt optimization baselines.} 
In Table~\ref{main result}, \ours consistently improves accuracy across all five reasoning tasks compared to manual prompting methods.
For instance, \ours achieves accuracy gains of 3.0\% (71.6\% → 74.6\%) on StrategyQA and 7.8\% (73.4\% → 81.2\%) on GSM8K compared to the best few-shot CoT baseline. 
When compared to zero-shot CoT prompting, \ours achieves even larger improvements, with accuracy rising by 8.8\% (65.8\% → 74.6\%) and 21.0\% (60.2\% → 81.2\%) on these two tasks. 
Additionally, \ours surpasses other automatic prompt optimization methods across four benchmarks. Specifically, it achieves a 1.4\% increase ($60.0\% \to 61.4\%$) on LogiQA 2.0, 1.2\% ($73.4\% \to 74.6\%$) on StrategyQA, 2.2\% ($79.0\% \to 81.2\%$) on GSM8K, and 1.9\% ($76.3\% \to 78.2\%$) on Date Understanding, while attaining suboptimal performance on Object Counting with an accuracy of 86.9\%.
Beyond per-task gains, \ours also achieves the highest weighted average accuracy across all datasets, as reported in Appendix~\ref{appendix:waa}, exceeding the best-performing baseline by 2.7\%.
These results highlight \ours's capability to generate higher-quality prompts, enabling more effective learning with limited data and outperforming all baseline methods.

\begin{table}
\centering
\resizebox{\columnwidth}{!}{
\begin{tabular}{ll}
\toprule
\bf Method        & \bf Accuracy (\%) \\
\midrule
\ours 
    & \textbf{81.2} \textcolor{gray}{\small \textpm 1.2}   \\
\hspace{0.1em} w/o \textsc{Text Residual Connection}
    & 68.8 \textcolor{gray}{\small \textpm 3.9} \daa{12.4} \\
\hspace{0.1em} w/o \textsc{\textsc{Perplexity-Informed Node Selection}}
    & 68.2 \textcolor{gray}{\small \textpm 1.8}  \daa{13.0} \\
\bottomrule
\end{tabular}
}
\caption{Ablation study results.
\textbf{Bold} value indicates the best performance.
Red arrows indicate the absolute performance decrease compared to the full model.}
\label{components}
\end{table}

\begin{table}
\centering
\resizebox{\columnwidth}{!}{
\begin{tabular}{lccc}
\toprule
\textbf{Metric} & \textbf{Perplexity} & \textbf{Entropy} & \textbf{Length} \\
\midrule
\textbf{Accuracy (\%)} & \textbf{81.2} \textcolor{gray}{\small \textpm 1.2} & 78.6 \textcolor{gray}{\small \textpm 1.0} \daa{2.6} & 73.6 \textcolor{gray}{\small \textpm 4.3} \daa{7.6} \\
\bottomrule
\end{tabular}
}
\caption{Comparison of different node selection metrics on GSM8K. \textbf{Bold} value indicates the best performance. Red arrows indicate the absolute performance decrease compared to perplexity.}
\label{tab:node_metrics}
\end{table}

\paragraph{Finding \#3: \ours demonstrates robust cross-task prompt optimization.} In Table \ref{main result}, \ours improves performance across all five tasks.
In contrast, other automatic prompt optimization methods show inconsistent generalization performance.
{
For instance, APE achieves only 57.4\% and 72.8\% accuracy on the LogiQA 2.0 and Date Understanding datasets, respectively, which are lower than the 59.0\% and 76.1\% achieved by zero-shot CoT prompting, indicating a degradation in performance.
}
Meanwhile, \ours demonstrates superior cross-task stability, successfully optimizing the initial prompt in all five tasks.

\begin{table}
\centering
\resizebox{.85\columnwidth}{!}{
\begin{tabular}{ll}
\toprule
\textbf{Embedding Model} & \textbf{Accuracy (\%)} \\
\midrule
\texttt{text-embedding-3-large} & \textbf{81.2} {\small \textcolor{gray}{\textpm 1.2}} \\
\texttt{text-embedding-3-small} & 77.8 {\small \textcolor{gray}{\textpm 2.7}} \daa{3.4} \\
\texttt{text-embedding-ada-002} & 60.2 {\small \textcolor{gray}{\textpm 1.5}} \daa{21.0} \\
\bottomrule
\end{tabular}
}
\caption{Impact of embedding model choice on performance. 
\textbf{Bold} value indicates the best performance.
Red arrows indicate the absolute performance decrease compared to the full model.
Our results show that more recent and larger models achieve better performance.
}
\label{tab:embedding}
\end{table}

\subsection{Further Analysis}
\paragraph{Hyperparametric sensitivity analysis of \ours.} On GSM8K, we systematically analyze the effect of tree width $K$, hyperparameters $b_1$, and $b_2$ on \ours, as visualized in Figure~\ref{figure:performance}. The results reveal the following key findings: (a) Increasing the tree width $W$ from 2 to 6 leads to a unimodal accuracy curve, initially rising and then declining, with peak accuracy of 81.2\% at $K=3$; (b) The parameter $b_1$ is negatively correlated with performance. As $b_1$ increases from 0.25 to 1, accuracy decreases by 11.2\%, underscoring the importance of dynamically removing redundant parent node information; (c) The parameter $b_2$ plays a critical role in balancing performance and computational overhead. As $b_2$ increases from 0 to 0.5, accuracy fluctuates non-monotonically, while the optimized prompt length rapidly decreases. Further increasing $b_2$ to 1 results in a decline in accuracy from 81.2\% to 73.4\%, with the optimized prompt length reduced to 26 tokens. The optimal trade-off between performance and efficiency is achieved at $b_2 = 0.5$, demonstrating that moderate length constraints can effectively eliminate redundant lexical units while preserving essential reasoning information.

\begin{figure*}[!th]
\centering
\includegraphics[width=\textwidth]{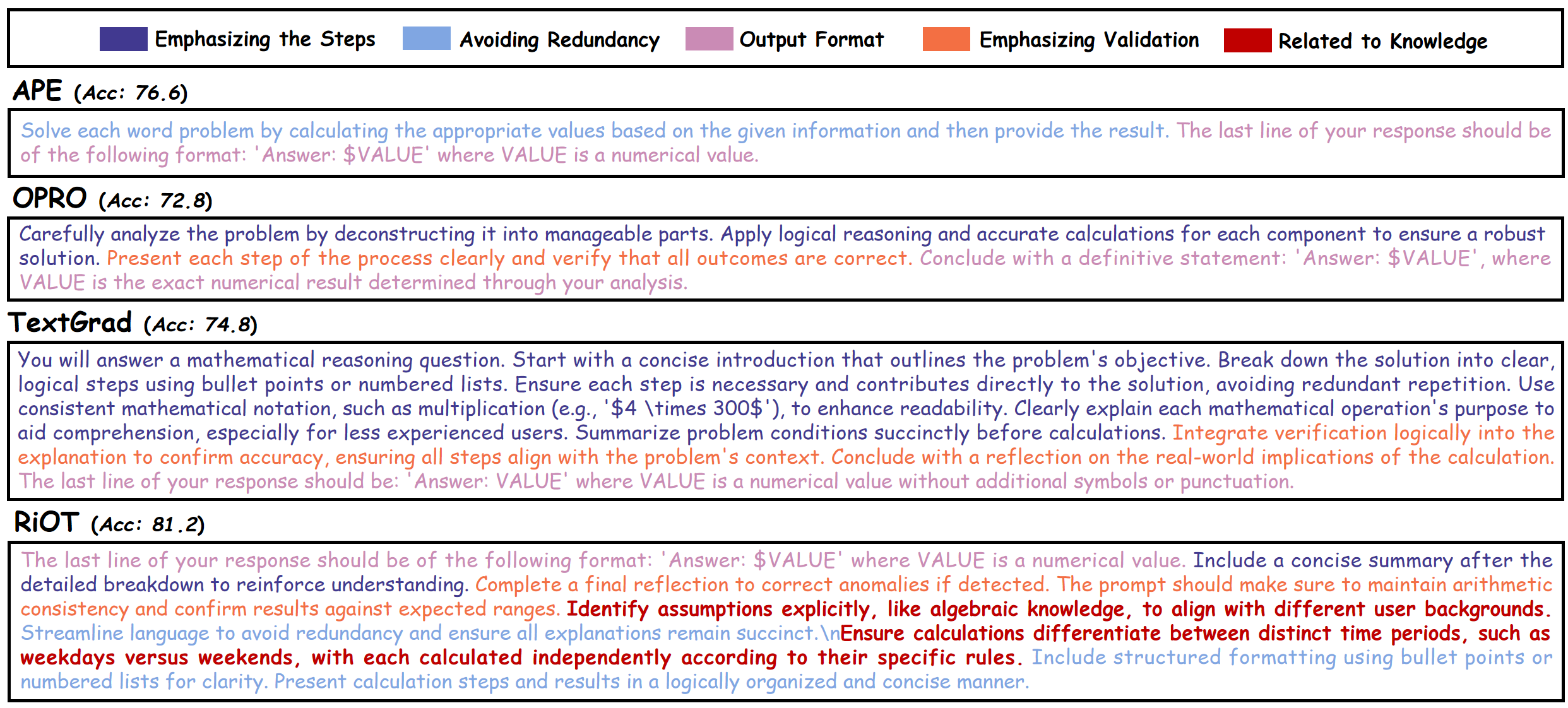}
\caption{List of optimized prompts on GSM8K by different prompt optimization methods, categorized according to five key dimensions: emphasizing the steps, avoiding redundancy, output format, emphasizing validation, and relating to domain-specific knowledge. \ours integrates all dimensions, enhancing reliability and aligning solutions with real-world contexts.}
\label{figure:case study}
\vspace{-0.45em}
\end{figure*}

\paragraph{Generalization analysis of \ours.} We conduct systematic evaluations to validate the generalization of \ours across two critical dimensions: \textbf{(1) Prompt Transferability}: Assessing whether prompts optimized for one model generalize to another. \textbf{(2) Model Transferability}: Evaluating \ours's ability to optimize prompts for distinct target models. In both experiments, prompts optimized by GPT-4o are evaluated on Gemini-1.5-flash~\cite{gemini1.5_report}, with the first experiment optimizing for GPT-3.5-turbo and the second for Gemini-1.5-flash. Additionally, we evaluate the zero-shot CoT prompt on Gemini-1.5-flash as the baseline. The results are shown in Table~\ref{generalization}. Optimizing prompts for GPT-3.5-turbo boosts the performance of Gemini-1.5-flash across three tasks. Furthermore, when Gemini-1.5-flash is used as the target model for prompt optimization with \ours, its performance surpasses the baseline across five tasks. For instance, on the StrategyQA dataset, accuracy improves by 1.8\% in the Prompt Transferability experiment and by 2.4\% in the Model Transferability experiment, both compared to the baseline. These results demonstrate the effectiveness of \ours in optimizing prompts for both model and prompt transferability.

\paragraph{Ablation study on each component of \ours on model performance.} In Table~\ref{components}, we evaluate individual component of \ours on GSM8K. In particular, we compare two variants: {\textbf{(1) w/o Text Residual Connection}: The optimal child node in each iteration is used as the parent node for the next, without incorporating any information from the current prompt, \textbf{(2) w/o Perplexity-Informed Node Selection}: Each optimization step generates only a single candidate prompt (\ie $K = 1$), disabling the selection mechanism.}
When the number of candidate prompts is limited to one, the effect of Text Residual Connection is minimal, resulting in a 13\% decrease in accuracy, highlighting the importance of diversity in automatic prompt optimization. Additionally, when there is a lack of inheritance between parent and child nodes during iterations, accuracy drops by 12.4\%, reinforcing our hypothesis that semantic drift occurs during the optimization process. {These results show that \ours benefits from both prompt diversity and sentence-level semantic fusion, and that both components are crucial for stable and effective optimization.
}

{\paragraph{Impact of node selection metrics.} To evaluate the impact of different node selection strategies in \ours, we compare three metrics: perplexity, entropy, and length, which capture different aspects of prompt quality—semantic likelihood, prediction uncertainty, and verbosity. As shown in Table~\ref{tab:node_metrics}, experimental results on GSM8K demonstrate that perplexity-informed selection consistently outperforms the other two, indicating its effectiveness in capturing semantic diversity and informativeness during prompt optimization. Formal definitions of entropy and length are provided in Appendix~\ref{appendix:metrics_definition}.}

\paragraph{Better embedding models lead to improved performance.} Table~\ref{tab:embedding} shows \ours's sensitivity to embedding quality on GSM8K, comparing distinct embedding models: text-embedding-3-large (top-tier)~\cite{openai2023textembedding}, text-embedding-3-small (mid-tier)~\cite{openai2023textembedding}, and text-embedding-ada-002 (base)~\cite{text-embedding-ada-002}. Accuracy drops by 3.4\% and 21\% respectively when downgrading models, revealing \ours's strong dependency on semantic discrimination capability. This dependency stems from the fact that lower-performing embedding models inadequately discriminate subtle lexical variations, impairing the identification of semantically similar units between parent and child nodes during residual connection. Consequently, suboptimal embedding models require careful calibration of the similarity threshold parameters $b_1$ and $b_2$ to mitigate performance loss.

\subsection{Case Study}
As shown in Figure~\ref{figure:case study}, we compare prompts optimized by different automatic prompt optimization methods on GSM8K. This comparison allows us to further explore \ours. Unlike baselines, \ours's optimized version uniquely combines verification mechanisms and domain adaptation, enforcing arithmetic consistency while eliminating redundant reasoning steps. Crucially, the explicit incorporation of temporal rule decoupling and algebraic assumption specification highlights \ours's ability to dynamically assimilate domain-specific patterns from training samples during optimization. This practical insight enables \ours to produce solutions with higher real-world alignment.
\section{Conclusion}
{
In this paper, we propose Residual Optimization Tree (\ours), a novel tree-based optimization framework that automatically generates higher-quality prompts. We employ a perplexity-informed node selection strategy to increase diversity in the prompt optimization process and incorporate a text residual connection to preserve the crucial semantic information. Experimental results show that \ours demonstrates substantial improvements in performance across five reasoning tasks while maintaining generalization.
}

\section*{Limitations}
Despite the promising results, our study has two main limitations, which reflect broader challenges in gradient-free automatic prompt optimization. First, our research focuses on textual tasks. However, the growing use of multimodal large language models (MLLMs) in real-world applications—ranging from visual question answering to robotic instruction synthesis—highlights the critical need for multimodal prompt engineering. Whether our method retains its effectiveness in the multimodal domain is an important direction for future exploration.
Second, we observe notable performance variation across different task categories, suggesting challenges in cross-task generalization for automatic prompt optimization. This issue is likely due to inherent differences in how LLMs organize and activate task-specific knowledge, pointing to the need for further investigation into the latent knowledge structures within LLMs.

\section*{Acknowledgements}
This work is funded by National Natural Science Foundation of China (U23A20496, 62302433), Zhejiang Provincial Key Research and Development Project of China (2024C01135), Zhejiang Provincial Natural Science Foundation of China (LQ24F020007) and the Ningbo Natural Science Foundation (2024J020). This work was supported by Ant Group.
\bibliography{custom}

\clearpage
\appendix

\section{Weighted Average Accuracy Across Datasets}
\label{appendix:waa}
To provide a holistic evaluation of each method's performance across all benchmark datasets, we report their Weighted Average Accuracy (WAA). The WAA is defined as:
\begin{equation}
\text{WAA} = \frac{\sum_{i=1}^{n} N_i \cdot A_i}{\sum_{i=1}^{n} N_i}
\end{equation}
where $N_i$ denotes the number of test samples and $A_i$ denotes the average accuracy on the $i$-th dataset. $n$ is the number of datasets.

As shown in Figure~\ref{fig:waa}, \ours achieves the highest WAA of 77.2\%, outperforming all baselines. Specifically, it exceeds the Zero-shot CoT method by 8.2\% and shows consistent advantages over automatic prompt optimization baselines including APE (5.5\%), OPRO (4.7\%), TextGrad (4.0\%), and DSPy (2.7\%).

\begin{figure}[H]
    \centering
    \includegraphics[width=\columnwidth]{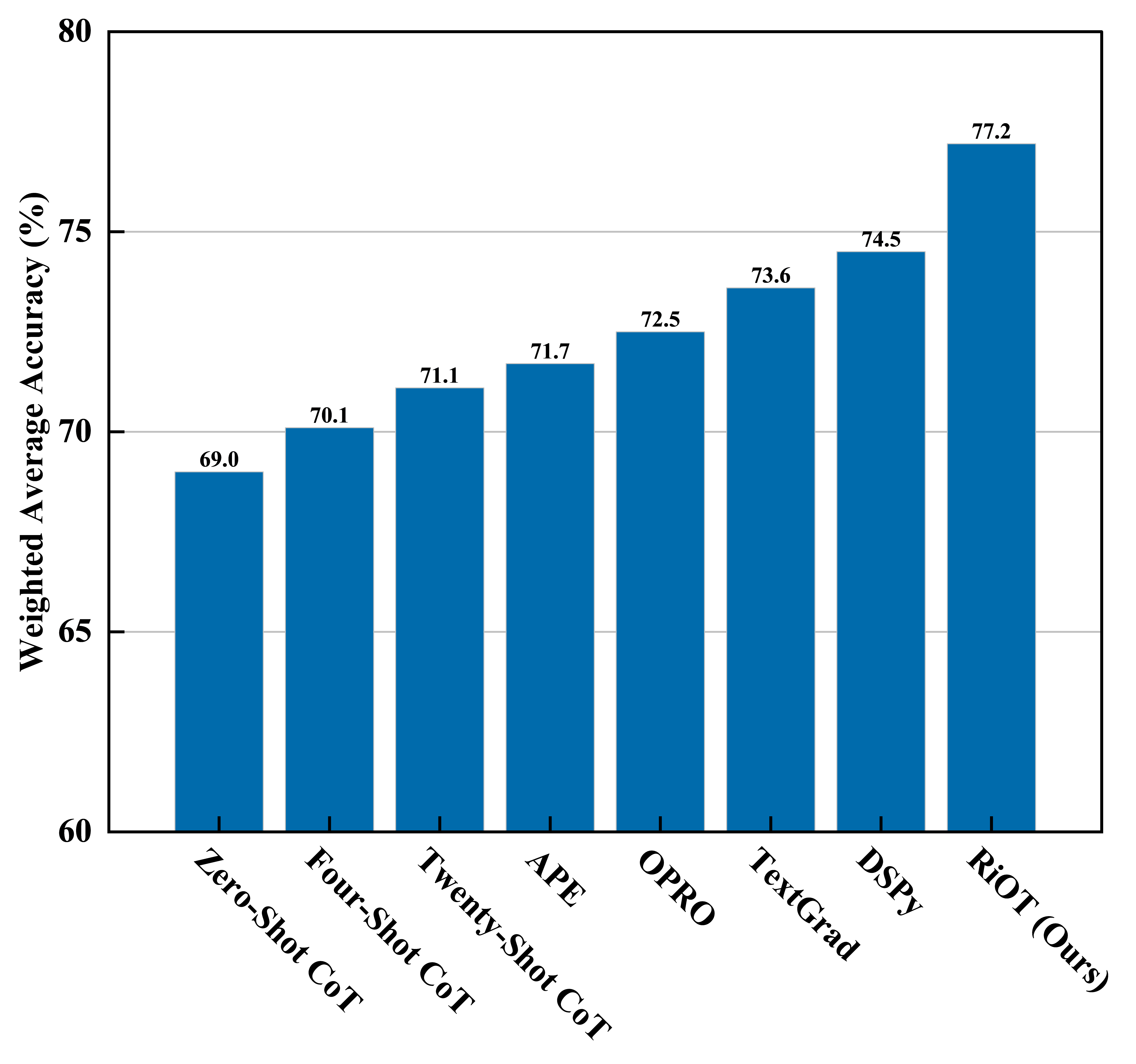}
    \caption{Weighted Average Accuracy (WAA) of different methods across all datasets. \ours achieves the highest WAA of 77.2\%, outperforming all baseline methods.
}
    \label{fig:waa}
\end{figure}

\section{Alternative Node Selection Metrics}
\label{appendix:metrics_definition}
We consider two alternative metrics to perplexity for scoring candidate prompts: Entropy and Length. Given a prompt sequence $(x_0,x_1, \ldots, x_J)$, the metrics are defined as:
\[
\text{Entropy} = -\sum_{j=1}^Jp_\theta(x_j \mid x_{<j})\log p_\theta(x_j\mid x_{<j})
\]
\[
\text{Length}=J
\]
where $p_\theta(x_j \mid x_{<j})$ denotes the probability assigned to token $x_j$ given its preceding context.

\section{Computational Cost Analysis}
In this section, we conduct an explicit analysis of the computational overhead introduced by \ours. Compared to TextGrad, the additional cost mainly stems from the need to generate $K$ candidate prompts at each optimization step (with $K = 3$ in our experiments) and to perform semantic fusion, which involves computing embeddings and evaluating semantic similarity for residual merging. 

Since \ours adopts TextGrad as its optimization backbone, we compare their runtime to assess the computational complexity. Theoretically, RiOT requires approximately $K$ times the optimization time of TextGrad. In practice, however, this overhead is effectively mitigated by leveraging multithreading, which enables parallel execution of candidate generation steps.

As shown in Table~\ref{tab:runtime}, the multithreaded version of \ours only incurs a modest increase in runtime (approximately 9\%) compared to TextGrad, while achieving a notable improvement in accuracy. This demonstrates that the trade-off between performance gain and computational cost is both reasonable and practical for real-world applications.

\begin{table}
\centering
\resizebox{\columnwidth}{!}{ % 控制表格宽度
\begin{tabular}{lccc}
\toprule
\textbf{Method} & \textbf{TextGrad} & \textbf{RiOT (Single)} & \textbf{RiOT (Multi)} \\
\midrule
\textbf{Time (s)} & 989.47 & 2741.41 & 1080.56 \\
\bottomrule
\end{tabular}
}
\caption{Runtime comparison between TextGrad and RiOT under different threading settings.}
\label{tab:runtime}
\end{table}

\section{Additional Evaluation on a Small-Scale, High-Difficulty Benchmark}
To further evaluate the generalization ability of \ours in complex reasoning scenarios, we conduct an additional experiment on the AMC12 benchmark. AMC12 (American Mathematics Competition 12) is a small-scale but high-difficulty dataset consisting of 83 math problems from the official 2022 and 2023 exams. It requires symbolic manipulation and multi-step reasoning, posing substantial challenges for prompt optimization methods. The dataset is split into training, validation, and test sets in a 20/20/43 ratio. As shown in Table~\ref{tab:amc12}, \ours achieves an accuracy of 46.0\%, surpassing all baselines and improving over the zero-shot CoT baseline by +5.1\%. These results serve as a further validation of \ours's robustness and effectiveness in low-resource, high-complexity settings.

\begin{table*}
\centering
\resizebox{\textwidth}{!}{
\begin{tabular}{lcccccccc}
\toprule
\textbf{Method} & Zero-Shot CoT & Four-Shot CoT & Twenty-Shot CoT & APE & OPRO & TextGrad & DSPy & \textbf{RiOT (Ours)} \\
\midrule
\textbf{Accuracy (\%)} & 40.9 & 37.2 & 30.7 & 42.8 & 42.8 & 40.9 & 41.9 & \textbf{46.0} \\
\textbf{$\Delta$} & +0.0 & -3.7 & -10.2 & +1.9 & +1.9 & +0.0 & +1.0 & \textbf{+5.1} \\
\bottomrule
\end{tabular}
}
\caption{Comparion of Automatic Prompt Optimization Methods on the AMC12 benchmark.}
\label{tab:amc12}
\end{table*}

\section{Details of Evaluation Datasets}
In this section, we provide detailed information about the five reasoning datasets used in our study. Table~\ref{tab:dataset distribution} summarizes dataset statistics, while Table~\ref{tab:task} provides representative examples. To ensure a fair comparison, all experiments follow consistent data sampling protocols: fixed-size development sets with 20 training and 20 validation samples each task. Below we detail each dataset’s unique characteristics.

\textbf{LogiQA 2.0}~\cite{logiqa} is a complex logical reasoning dataset derived from the Chinese Civil Service Examination. We use its nature language inference subset, which includes 6,871 training samples, 3,807 validation samples, and 3,240 test samples. The problems are categorized into five types: categorical reasoning, sufficient conditional reasoning, necessary conditional reasoning, disjunctive reasoning, and conjunctive reasoning. We sample 160 test samples that cover these five reasoning types from the original dataset.

\textbf{StrategyQA}~\cite{StrategyQA} is a commonsense reasoning task that evaluates the model's ability to infer unstated intermediate facts and strategically combine them to derive correct answers, leveraging general world knowledge. The original dataset consists of 2,061 training samples and 229 test samples. Following previous research~\cite{Reconcile, textgrad, llm_debate, ToT}, we sample 100 test samples for evaluation.

\textbf{Object Counting}~\cite{Big-Bench, BBH} measures the model's ability to count objects within a given context while ignoring distractors. This task emphasizes fundamental reasoning over arithmetic by focusing on semantic understanding. The original dataset contains 250 test samples. For our study, we sample 210 test samples for evaluation.

\textbf{GSM8K}~\cite{GSM8K} evaluates the model's ability to solve basic mathematical problems requiring multi-step reasoning. This task focuses on contextual understanding and numerical computation. The original dataset consists of 7,473 training samples and 1,319 test samples. Following previous research~\cite{Reconcile, textgrad, llm_debate, ToT}, we sample 100 test samples for evaluation.

\textbf{Date Understanding}~\cite{Big-Bench, BBH} assesses the model's ability to comprehend temporal information, requiring reasoning over explicit or implicit data contexts to identify specific dates. The original dataset contains 369 test samples. For our study, For our study, we sample 329 test samples for evaluation.

% \begin{itemize}
% \item \textbf{Object Counting}~\cite{Big-Bench, BBH} measures the model's capacity to count objects within a given context while ignoring distractors. This task emphasizes fundamental reasoning over arithmetic by focusing on semantic understanding.
% \item \textbf{GSM8K}~\cite{GSM8K} evaluates the model's ability to solve basic mathematical problems requiring multi-step reasoning. This task focuses on contextual understanding and numerical computation.
% \item \textbf{Date Understanding}~\cite{Big-Bench, BBH} assesses the model's ability to comprehend temporal information, requiring reasoning over explicit or implicit data contexts to identify specific dates.
% \end{itemize}
\label{appendix: dataset}
\begin{table}[h!]
\centering
\resizebox{\columnwidth}{!}{
\begin{tabular}{cccc}
\toprule  % 表格顶部的横线
\textbf{Task} & \textbf{Training} & \textbf{Validation} & \textbf{Test} \\
\hline
LogiQA 2.0  & 20 & 20 & 160 \\
StrategyQA  & 20 & 20  & 100 \\
Object Counting & 20 & 20 & 210 \\
GSM8K & 20 & 20 & 100 \\
Date Understanding & 20 & 20 & 329 \\
\bottomrule  % 表格底部的横线
\end{tabular}
}
\caption{Dataset Statistics.}
\label{tab:dataset distribution}
\end{table}

\section{Details of Manual Prompt Construction}
\label{appendix: manual prompt}
In this section, we provide detailed information about the design and implementation of the zero-shot and few-shot CoT prompts used as baselines in our experiments. Both prompt variants follow a structured template to ensure consistency across tasks while accommodating task-specific requirements. The general template $\mathcal{P}_{\text{CoT}}^k$ is defined as:
\begin{equation}
    \mathcal{P}_{\text{CoT}}^k = \mathcal{P}_{\text{task}} + \mathcal{P}_{\text{think}} + \mathcal{P}_{\text{format}} + \{\mathcal{P}_{\text{demo}}^{(i)}\}_{i=1}^k
\end{equation}
where $\mathcal{P}_{\text{task}}$ defines the precise task definition and the objective, $\mathcal{P}_{\text{think}}$ is the reasoning guidance, $\mathcal{P}_{\text{format}}$ denotes output structure constraints and $\{\mathcal{P}_{\text{demo}}^{(i)}\}_{i=1}^k$ is the demonstration examples. The zero-shot CoT prompts $\{\mathcal{P}_{\text{CoT}}^0\}$ is exemplified in Table~\ref{tab:prompts}. The few-shot CoT prompts $\{\mathcal{P}_{\text{CoT}}^k\}$ extend this template by appending demonstrations from the training set, and each demonstration consists of an input-output pair. An example is visualized in Figure~\ref{figure:4-shot}. These manual prompts represent the most widely adopted approaches in practice and serve as fundamental baselines for evaluating prompt engineering techniques.

\begin{table*}[ht]
\centering
\resizebox{\textwidth}{!}{
\begin{tabular}{lccccc}
\toprule
           & \multicolumn{2}{c}{\bf \textsc{True / False}} & \multicolumn{2}{c}{\bf \textsc{Generative}} & \bf \textsc{Multiple-choice} \\
           \cmidrule(lr){2-3}                          \cmidrule(lr){4-5}                        \cmidrule(lr){6-6}
\bf Implementation & LogiQA 2.0 & StrategyQA & Object Counting & GSM8K & Date Understanding \\
           & (N=160) & (N=100) & (N=210) & (N=100) & (N=329) \\
\midrule
Sentence-based 
                & 61.4 \textcolor{gray}{\tiny \textpm 1.5}          
                & 74.6 \textcolor{gray}{\tiny \textpm 1.4}       
                & 86.9 \textcolor{gray}{\tiny \textpm 0.6}
                & 81.2 \textcolor{gray}{\tiny \textpm 1.2}      
                & 78.2 \textcolor{gray}{\tiny \textpm 0.6}               
                \\
LLM-based
                & 60.1 \textcolor{gray}{\tiny \textpm 1.4} \daa{1.3}
                & 72.4 \textcolor{gray}{\tiny \textpm 1.2} \daa{2.2}
                & 81.7 \textcolor{gray}{\tiny \textpm 1.3} \daa{5.2}
                & 75.4 \textcolor{gray}{\tiny \textpm 2.8} \daa{5.8}
                & 80.2 \textcolor{gray}{\tiny \textpm 0.3} \uaa{2.0}
                \\
\bottomrule
\end{tabular}
}
\caption{Comparison of performance between two implementations for text residual connection across five tasks. We report accuracy (\%) along with standard deviation. Red arrows indicate the absolute performance decrease, while green arrows indicate the absolute performance increase, both compared to the sentence-based implementation. $N$ represents the number of samples used for testing. The sentence-based implementation outperforms the LLM-based implementation.}
\label{tab:distinct residual}
\end{table*}
\section{An Alternative Implementation of the Text Residual Connection}
In our exploration of Text Residual Connection, we experiment with two distinct implementation approaches. The first approach, introduced in the Section~\ref{sec:residual learning}, is the \textbf{sentence-based} Implementation, which utilizes sentence tokenization and vector similarity to perform the text residual connection. The second approach, the \textbf{LLM-based} Implementation, leverages the strong meta-reasoning capabilities of modern LLMs without explicit lexical analysis. To be specific, we develop a rigorously engineered prompt template to instruct LLMs in performing text residual connection. As shown in Figure~\ref{figure:end2end}, the prompt template incorporates four key principles: \textbf{(1) Task Specification}: Explicit definition of input-output relationships. \textbf{(2) Constraint Enumeration}: Clear operational boundaries. \textbf{(3) Exemplar Guidance}: Illustrative examples demonstrating ideal behavior. \textbf{(4) Error Prevention}: Anticipating and mitigating common failure modes.

We evaluate the performance of these two implementations across five datasets, using GPT-4o to implement the LLM-based Implementation. The results, presented in Table~\ref{tab:distinct residual}, show that the sentence-based implementation generally outperforms the LLM-based implementation, particularly in tasks involving binary classification (True/False) and generative tasks. Additionally, during the experimental process, we observe that the LLM-based implementation does not always strictly adhere to the task requirements in the same way as the sentence-based implementation. Specifically, the LLMs tend to modify or omit parts of the original content in the prompt during execution. For these reasons, we choose to use the sentence-based implementation for the main experiments in this study, as it provide more consistent and reliable results.

Despite these challenges, the LLM-based implementation remains promising, especially given its end-to-end nature and the lack of the need for hyperparameter tuning. This characteristic makes it easier to integrate directly into existing optimization frameworks, which is a substantial advantage in practical applications. Further development and exploration are required to refine this approach.

\begin{table*}[h!]
\centering
\begin{tabular}{>{\centering\arraybackslash}m{3cm} >{\arraybackslash}m{12cm}}
\toprule  % 表格顶部的横线
\multicolumn{1}{c}{\textbf{Task}} & \multicolumn{1}{c}{\textbf{Zero-Shot Prompt (Initial Prompt)}}  \\  % 表头内容居中
\hline  % 表头下方的横线
LogiQA 2.0 & \textcolor[RGB]{65,57,144}{You will solve logical reasoning problems based on the given facts.} \textcolor[RGB]{192,0,0}{Think step by step.} \textcolor[RGB]{128,166,226}{The last line of your response should be of the following format: "Answer: YES" or "Answer: NO".}
 \\
 \\
StrategyQA  & \textcolor[RGB]{65,57,144}{You will answer a commonsense reasoning task.} \textcolor[RGB]{192,0,0}{Think step by step.} \textcolor[RGB]{128,166,226}{The last line of your response should be of the following format: "Answer: YES" or "Answer: NO".}
 \\
 \\
Object Counting  & \textcolor[RGB]{65,57,144}{You will answer a reasoning question.} \textcolor[RGB]{192,0,0}{Think step by step.} \textcolor[RGB]{128,166,226}{The last line of your response should be of the following format: `Answer: \$VALUE' where VALUE is a numerical value.}
 \\
 \\
GSM8K & \textcolor[RGB]{65,57,144}{You will answer a mathematical reasoning question.} \textcolor[RGB]{192,0,0}{Think step by step.} \textcolor[RGB]{128,166,226}{The last line of your response should be of the following format: `Answer: \$VALUE' where VALUE is a numerical value.}
 \\
 \\
Date Understanding & \textcolor[RGB]{65,57,144}{You will answer a multiple-choice question related to date understanding.} \textcolor[RGB]{192,0,0}{Think step by step.} \textcolor[RGB]{128,166,226}{The last line of your response should be of the following format: `Answer: \$VALUE' where VALUE is a single letter.}
 \\
\bottomrule  % 表格底部的横线
\end{tabular}
\caption{Zero-Shot Prompts (initial prompt for optimization) for various reasoning tasks. Each prompt consists of three components: \textcolor[RGB]{65,57,144}{the task definition $\mathcal{P}_\text{task}$}, \textcolor[RGB]{192,0,0}{the reasoning guidance $\mathcal{P}_\text{think}$}, and \textcolor[RGB]{128,166,226}{the output structure constraints $\mathcal{P}_\text{format}$}.}
\label{tab:prompts}
\end{table*}

\begin{table*}[h!]
\centering
\begin{tabular}{>{\centering\arraybackslash}m{3cm} >{\arraybackslash}m{12cm}}
\toprule  % 表格顶部的横线
\multicolumn{1}{c}{\textbf{Task}} & \multicolumn{1}{c}{\textbf{Example Problem}}\\  % 表头内容居中
\hline  % 表头下方的横线
LogiQA 2.0 & \textbf{Input}: \textit{Given the fact}: Balance is particularly important when (...) If all the media were to adopt such a perverse interpretation of balanced reporting, the public would be given a picture of a world where each party in every conflict had an equal measure of justice on its side, contrary to our experience of life and, indeed, our common sense.
\textit{Does it follow that}: The main point of the argument is that balanced reporting requires impartially revealing injustices where they occur no less than fairly presenting the views of each party in a conflict. \textbf{Ground Truth}: Yes.
 \\
 \\
StrategyQA  & \textbf{Input}: \textit{Question}: Are months based on the solar cycle? \textbf{Ground Truth}: Yes.
 \\
 \\
Object Counting  & \textbf{Input}: \textit{Question}: I have a fridge, an oven, a car, a toaster, a microwave, a table, and a bed. How many objects do I have? \textbf{Ground Truth}: 7.
 \\
 \\
GSM8K & \textbf{Input}: \textit{Question}: Stefan goes to a restaurant to eat dinner with his family. They order an appetizer that costs \$10 and 4 entrees that are\$20 each. If they tip 20\% of the total for the waiter, what is the total amount of money that they spend at the restaurant? \textbf{Ground Truth}: 108.
 \\
 \\
Date Understanding & \textbf{Input}: \textit{Question}: In the US, Thanksgiving is on the fourth Thursday of November. Today is the US Thanksgiving of 2001. What is the date today in MM/DD/YYYY? \textit{Choices}: A. 01/16/2003 B. 11/21/2002 C. 09/04/2002 D. 08/24/2002 E. 11/22/2002 F. 11/23/2002 \textbf{Ground Truth}: E.
 \\
\bottomrule  % 表格底部的横线
\end{tabular}
\caption{Example problems and their corresponding ground truths for various reasoning tasks. The tasks cover a range of domains, including logical reasoning (LogiQA 2.0), commonsense reasoning (StrategyQA), semantic understanding (Object Counting), mathematical reasoning (GSM8K), and temporal reasoning (Date Understanding).}
\label{tab:task}
\end{table*}

\begin{figure*}[h!]
\centering
\includegraphics[width=\textwidth]{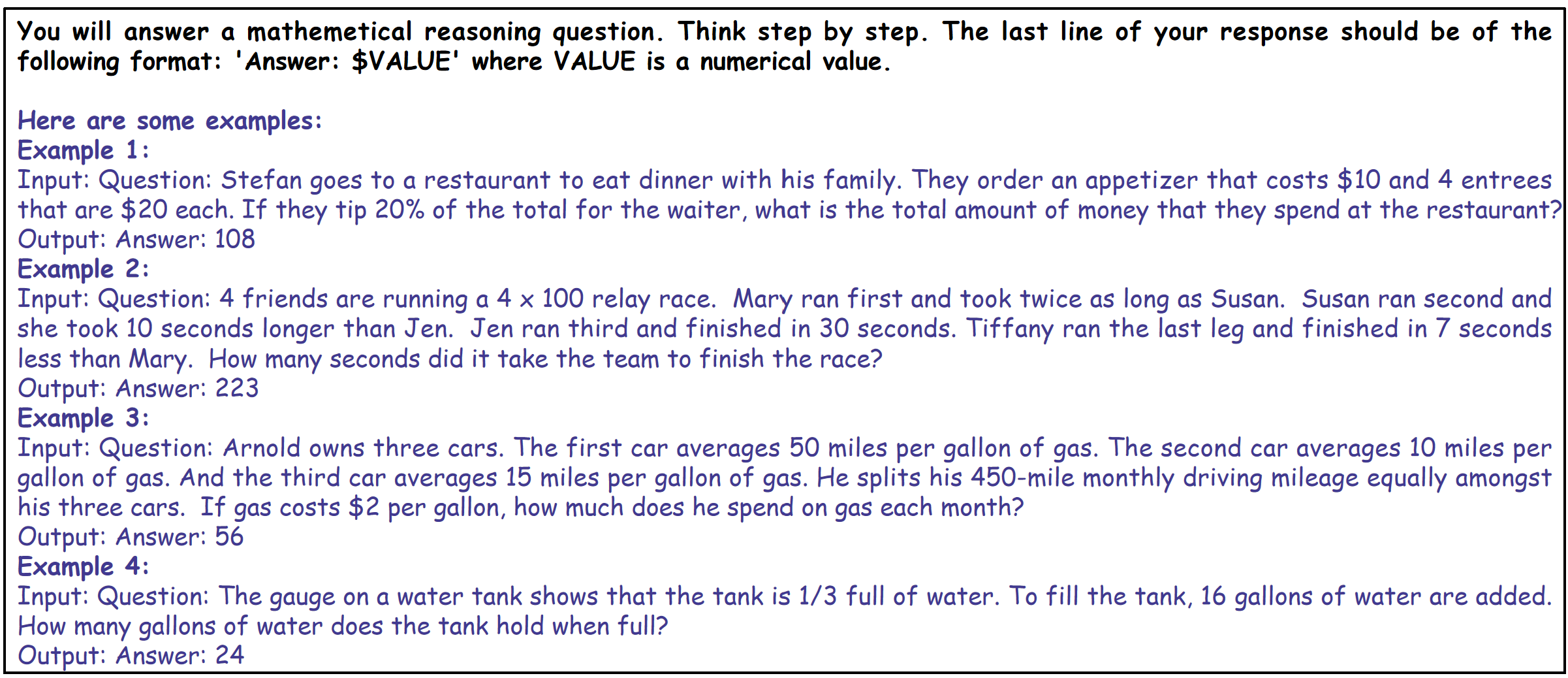}
\caption{An example of few-shot prompts from GSM8K. The purple section represent the \textcolor[RGB]{65,57,144}{demonstration examples ${\{\mathcal{P}_{\text{demo}}^{(i)}\}_{i=1}^k}$}. The examples are drawn from the training set.}
\label{figure:4-shot}
\end{figure*}

\begin{figure*}[h!]
\centering
\includegraphics[width=\textwidth]{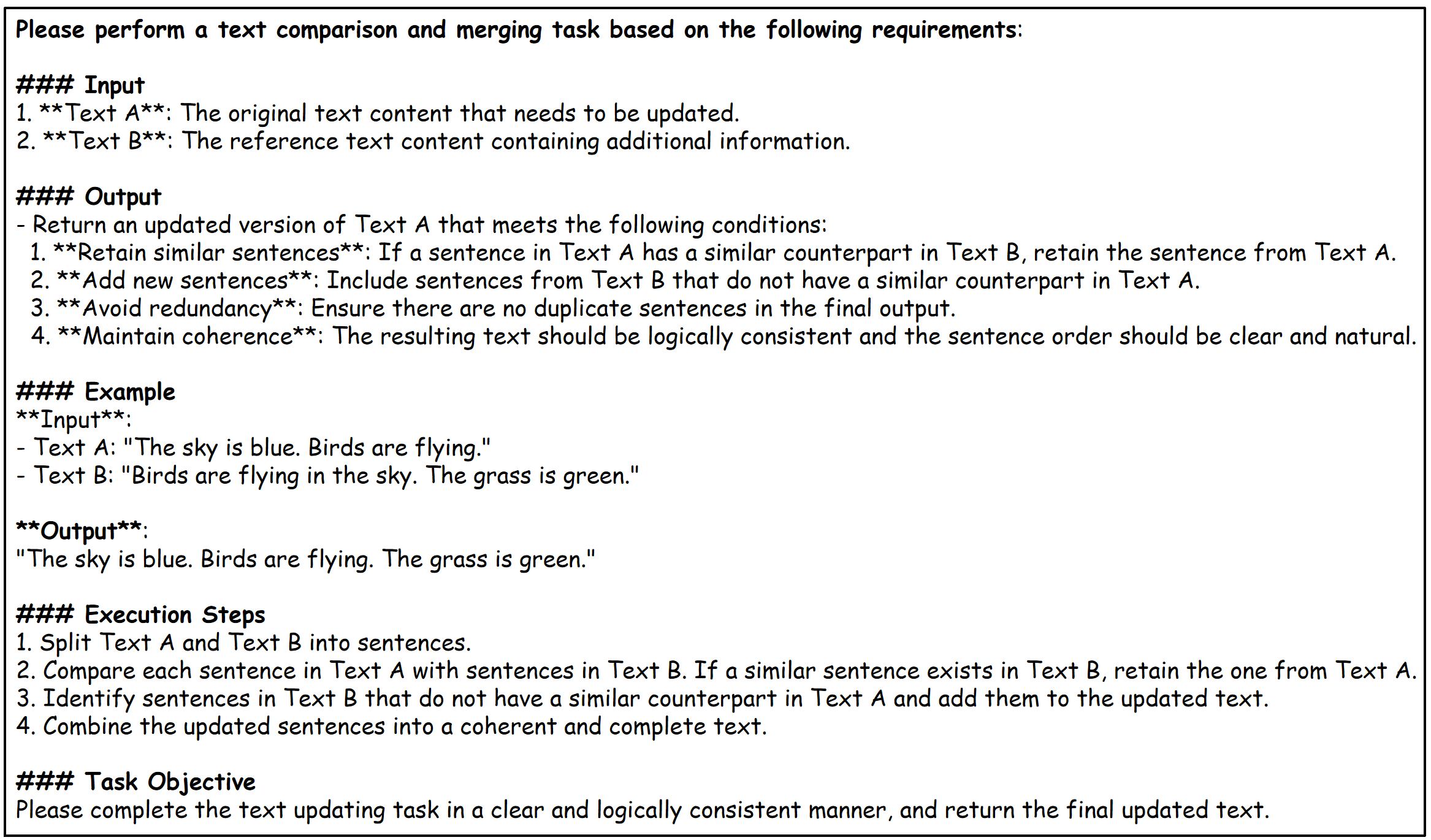}
\caption{The prompt template for the LLM-based Implementation of Text Residual Connection.}
\label{figure:end2end}
\end{figure*}

\end{document}